\documentclass{article} 
\usepackage{multirow}
\usepackage{floatflt}
\usepackage{hyperref}
\usepackage{graphicx}
\usepackage{amsmath, amsthm, amssymb}
\usepackage[margin=0.8in]{geometry}

\newcommand{\RR}{\mathcal{R}}
\newcommand{\OMIT}[1]{}

\newcommand{\del}[1]{}
\newcommand{\mcorr}[1]{{{#1}}}

\begin{document}

\title{A new protein binding pocket similarity measure based on comparison of 3D atom clouds: application to ligand prediction}
\author{Brice Hoffmann$^{*}$, Mikhail Zaslavskiy$^{*}$, Jean-Philippe Vert and V{\'e}ronique Stoven \\
Mines ParisTech, CBIO and CMM, France,\\ Institut Curie and INSERM U900, Paris F-75248 France.\\
 $^*$ Contributed equally
}

\maketitle
\begin{abstract}
\noindent \textbf{Motivation:} Prediction of ligands for proteins of known 3D structure is important to understand structure-function relationship, predict molecular function, or design new drugs.\\
\textbf{Results:}
 We explore a new approach for ligand prediction in which binding pockets are represented by atom clouds. Each target pocket is compared to an ensemble of pockets of known ligands. Pockets are aligned in 3D space with further use of convolution kernels between clouds of points. Performance of the new method for ligand prediction is compared to those of other available measures and to docking programs. We discuss two criteria to compare the quality of similarity measures: area under ROC curve (AUC) and classification based scores. We show that the latter is better suited to evaluate the methods with respect to ligand prediction. Our results on existing and new benchmarks indicate that the new method outperforms other approaches, including docking.\\
\textbf{Availability:} The new method is available at  \href{http://cbio.ensmp.fr/paris}{http://cbio.ensmp.fr/paris/}\\
\textbf{Contact:} \href{mikhail.zaslavskiy@mines-paristech.fr}{mikhail.zaslavskiy@mines-paristech.fr}
\end{abstract}
\section{Introduction} 
One of the main goals of structural biology is to  predict, from the 3D fold of a protein, its interacting partners, which in turn is related to its molecular function. However, understanding this structure-function relationship is still today an open question, and no reliable tool is available to permit such a prediction. Current efforts concentrate on local 3D approaches, focusing on identification and comparison of binding pockets, in order to predict the natural ligand for a protein, with the underlying idea that proteins sharing similar binding sites are expected to bind similar ligands. The same strategy also applies to the problem of identifying new drug precursors for a therapeutic target protein. 

The comparison of 3D binding pockets is an active field of research, and during the last  decade, many new methods were proposed. \cite{Morris2005Real} considered  a method based on using real spherical harmonic expansion coefficients, \cite{Gold2006SitesBase} used a specialized geometric hashing procedure as the core of the SitesBase web server, \cite{Shulman2008MultiBind} used multiple common point set detection method. An approach proposed by \cite{Schalon2008Simple} is based on a triangle-discretized sphere representation of  binding pockets. \cite{Weskamp2007Multiple} and \cite{Najmanovich2008Detection} considered graph-based representations of binding pockets and applied graph matching algorithms.

In this paper, we explore the potential of a new approach in which binding pockets are represented by clouds of atoms in 3D space  potentially baring additional labels such as partial charge or atom type. The new similarity measure is based on the alignment of protein pockets with further use of convolution kernel between 3D point clouds. We study how the proposed method may be used to predict a ligand for a given pocket by comparing it to a set of pockets with  known ligand. 

Here, we do not discuss the problem of pocket detection. In our experiments, we extracted pockets on the basis of known protein-ligand crystal structures as it was done by \cite{Kahraman2007Shape}. In cases where the binding site is unknown, various programs have been developed to locate depressions on protein surfaces and could be used to identify putative binding sites (\cite{Glaser2006Method}). 

An important question in this paper is the evaluation of pocket similarity measures. We discuss two criteria to compare the quality of similarity measures on the basis of their ability to detect pockets binding the same ligand: area under ROC curve (AUC) and classification based scores. We compare our method with some existing state of the art algorithms on different benchmark datasets. Since we evaluate methods for binding pocket comparison according to their ability to predict ligands, we also report the performance of docking methods, on the same benchmark datasets. Finally, we also discuss possible extensions of the proposed method to other applications such as protein function prediction or ligand comparison.

\section{Methods} 
\label{sec:methods} 
\subsection{Convolution kernel between atom clouds}
\label{sec:convolution}

In our model, a binding pocket is described by a set of atoms in 3D space. Our objective is to construct a  similarity measure between pockets, which may be used to identify pockets binding  the same ligand. 

Let $P=(x_i,l_i)_{i=1}^{N}$ denote a binding pocket consisting of $N$ atoms, where $x_i \in \RR^3$ is a 3D vector representing atom coordinates, and $l_i$ is a label (discrete or real valued) that may be used to bare additional information on the atoms (for example, atom type, atom partial charge, or amino acid type).

A classical approach for pocket comparison consists in iterative alignment of \mcorr{two pockets} and further counting of overlapping atoms, usually within a tolerance of 1\AA. Different implementations of this principle \del{is}  \mcorr{may be} found in such  methods as Tanimoto index \cite{Willett1986Implementation}, the SitesBase algorithm (Poisson index), or the MultiBind algorithm \cite{Shulman2008MultiBind}. The alignment is made to maximize the number of overlapping atoms, which is generally a good indicator of pocket similarity. 

However, atoms may have different positions but play equivalent roles in ligand binding, and the role of one atom in one pocket may be played by a group of atoms in another one. These observations lead us to the idea of an alternative smooth score which does not count the number of overlapping atoms, but rather uses a weighted number of atoms having closed positions. We first consider the case where labels are ignored, and only atom coordinates are used to measure the similarity between pockets, and then explain how the information on atom labels may be introduced in the new similarity measure.

Given two pockets $P_1$ and $P_2$ the similarity measure $K(P_1,P_2)$ is defined as follows
\begin{equation}
K(P_1,P_2)=\sum_{x_i\in P_1}\sum_{y_j\in P_2} e^\frac{-||x_i-y_j||^2}{2\sigma^2}.
\label{eq:3D_kernel}
\end{equation}    
This similarity measure  defines in fact a positive definite kernel, i.e. it may be considered as a true scalar product on the set of  atom clouds representing binding pockets \cite{Schoelkopf2004Kernel}. Implicitly, it  defines a distance between pockets:  $D(P_1,P_2)=K(P_1,P_1)+K(P_2,P_2)-2K(P_1,P_2)$ which has all standard properties of a true metric (non-negativity, identity of indiscernibles, symmetry, triangular inequality). The parameter $\sigma$ characterizes the sensitivity of the similarity measure (\ref{eq:3D_kernel}) to points relative displacements. When $\sigma$ is small, only atoms of two pockets which are very close to each other significantly contribute to  $K(P_1,P_2)$. On the contrary, when $\sigma$ is large, almost all pairs of atoms contribute to $K(P_1,P_2)$.  


The kernel (\ref{eq:3D_kernel}) is an example of a convolution kernel \cite{Haussler1999Convolution,Gartner2002Multi-Instance} between point sets. Alternative kernels may be constructed by substituting the Gaussian kernel  $e^\frac{-||x_i-y_j||^2}{2\sigma^2}$ by any other kernel between 3D vectors $x_i$ and $y_j$. 

Interestingly, the kernel (\ref{eq:3D_kernel})  may be seen as a particular case of kernel between point sets  defined as a kernel between distribution function estimated from point sets \cite{Kondor2003Kernel}. More precisely, let us represent each binding pocket $P_i$ by a distribution of masses defined as the sum of Gaussian with bandwidth $\sigma/\sqrt{2}$ functions centered on the pocket atoms, namely:$$ f_{P_i}(x)=\sum_{x_i\in P_i}e^{-\frac{||x-x_i||^2}{\sigma^2}}.$$
Then kernel (\ref{eq:3D_kernel}) between pockets $P_1$ and $P_2$ can be recovered, up to a scaling constant, as the  scalar product in $L_2(\RR^3)$ between the associated distributions because:
\OMIT{\begin{equation}
\begin{split}
&K(P_i,P_j)=K(f_{P_i},f_{P_j})=\langle f_{P_1},f_{P_2}\rangle_{L_2(R^3)}=\\
&C_1\int_{R^3}\sum_{x_i\in P_i}e^{-\frac{||x-x_i||^2}{2\sigma^2}}\sum_{y_i\in P_j}e^{-\frac{||x-y_j||^2}{2\sigma^2}}dx=C_1\sum_{x_i,y_j}\int_{R^3}e^{-\frac{||x-x_i||^2}{2\sigma^2}} e^{-\frac{||x-y_j||^2}{2\sigma^2}}=\\
&\qquad \qquad \qquad \qquad \qquad \qquad C_2\sum_{x_i\in P_1}\sum_{y_j\in P_2} e^{-\frac{-||x_i-y_j||^2}{\sigma^2}},
\end{split}
\end{equation*}
\noindent where $C_1$ and $C_2$ are some positive constants.   
}
{\begin{equation*}
\begin{split}
\langle f_{P_1},f_{P_2}\rangle_{L_2(\RR^3)}
=\int_{\RR^3}\sum_{x_i\in P_i}e^{-\frac{||x-x_i||^2}{\sigma^2}}\sum_{y_i\in P_j}e^{-\frac{||x-y_j||^2}{\sigma^2}}dx\\
= \sum_{\substack{x_i\in P_1\\ y_j\in P_2}} \int_{\RR^3}e^{-\frac{||x-x_i||^2}{\sigma^2}} e^{-\frac{||x-y_j||^2}{\sigma^2}}dx=C \sum_{\substack{x_i\in P_1\\ y_j\in P_2}} e^{-\frac{-||x_i-y_j||^2}{2\sigma^2}}=CK(P_1,P_2)\,,
\end{split}
\end{equation*}
where $C$ is a positive constant.

However, formula (\ref{eq:3D_kernel}) is not fully appropriate in practice, because the proposed measure is not invariant upon rotations and translations of the binding pockets. Therefore, we define a similarity measure {\it sup-CK} as the maximum of (\ref{eq:3D_kernel}) over all possible rotations and translations of one of the two pockets: 

\begin{equation} 
\mbox{sup-CK}(P_1,P_2)=\max_{R,y_t}\sum_{x_i\in P_1,\\y_j\in P_2} e^\frac{||x_i-(Ry_j+y_t)||^2}{2\sigma^2},
\label{eq:3D_kernel_max}
\end{equation}
\noindent where $R$ is an orthonormal rotation matrix and $y_t$ is a translation vector. {\it Sup-CK} is not a positive definite measure anymore, but it can still be used as a similarity score.  Furthermore,  to evaluate {\it sup-CK}, we now need to maximize a non-concave function over the set of rotations and translations, which may have many local maxima. Exact maximization of this non-concave function is a hard optimization problem and we propose to estimate an approximate solution by running a gradient ascent algorithm, starting from many different initial points, and taking the best local maximum. The optimization algorithm may be significantly accelerated by choosing an initial point close to the global optimum. In the case of binding pockets, a good approximation of the  optimal translation vector $y_t$ is the vector which translates the geometric center of $P_2$ into the geometric center of $P_1$, 
$y_t=\frac{1}{N_1}\sum_{x_i\in P_1}x_i-\frac{1}{N_2}\sum_{y_i\in P_2}y_i.$
The approximated rotation matrix $R$ superposes the first principal axis of $P_2$ with the first principal axis of $P_1$,  the second one with the second one, and the third one with the third one. Since principal vectors are defined up to a sign, the  two signs for all principal vectors of one of the binding pockets have to be tested (there are $2^3$ combinations). If some of the pocket axes have close lengths, then it may be also interesting to consider rotations which superpose the first principal axis of one pocket with the second principal axis of the other one.

Gradient ascent method requires to calculate the gradient of the function in (\ref{eq:3D_kernel_max}) with respect to $R$ and $y_t$. Calculation of the gradient components related to $y_t$ is straightforward:
$$
\nabla_{y_t}=\frac{1}{\sigma^2}\sum_{x_i\in P_1,\\y_j\in P_2}(x_i-(Ry_j+y_t))e^\frac{||x_i-(Ry_j+y_t)||^2}{2\sigma^2}.
$$

Since the set of rotation matrices is a 3D manifold embedded in 9D space, we cannot take derivatives with respect to each element of matrix $R$. Instead, we use the Euler representation of the rotation matrix:
\begin{equation}
\begin{split}
R=R_XR_YR_Z=
 \left[\begin{array}{ccc} 1 & 0 & 0\\ 0 & \cos\phi & \sin\phi\\ 0 & -\sin\phi & \cos\phi \end{array} \right] \left[\begin{array}{ccc} \cos\theta & 0 & -\sin\theta\\ 0 & 1 & 0\\ \sin\theta & 0 & \cos\theta \end{array} \right]\left[\begin{array}{ccc} \cos\psi & \sin\psi & 0\\ -\sin\psi & \cos\psi & 0\\ 0 & 0 & 1 \end{array} \right]\,,
\end{split}
\end{equation}
where $R$ is expressed as a function of $(\phi,\theta,\psi)\in[0;2\pi)^3$. The derivatives of the maximand in (\ref{eq:3D_kernel_max}) are now calculated with respect to $(\phi,\theta,\psi)$, for instance,
\begin{equation*}
\begin{split}
\nabla_{\theta}=\frac{1}{\sigma^2}\sum_{\substack{x_i\in P_1\\ y_j\in P_2}}e^\frac{||x_i-(Ry_j+y_t)||^2}{2\sigma^2}(x_i-(Ry_j+y_t))^T (R_X\frac{\partial R_Y}{\partial \theta}R_Z y_j)\,.
\end{split}
\end{equation*}

As mentioned above, it may be interesting to use additional information on binding pocket atoms (such as atom type or charge). Let us suppose that this information is represented by labels $l_i$ (which may be discrete or real variables, or multidimensional vectors) with an associated similarity measure. For example, to measure the similarity between categorical labels like atom types, the Dirac function $1_{l_i=l_j}$ may be used. In our experiments, we  use atom partial charges as atom labels, with a Gaussian kernel  $K_{\mbox{L}}(l_i,l_j)=e^{-\frac{(l_i-l_j)^2}{\lambda}}$. Of course, other similarity measures may be used as well.

Finally, atom labels are used to re-weight the contribution of two atoms $x_i$ and $y_j$ by $K_{\mbox{L}}(l_i,l_j)$ in (\ref{eq:3D_kernel_max}):
\begin{equation}
\mbox{sup-CK}_L(P_1,P_2)=\max_{R,y_t}\sum_{\substack{x_i\in P_1\\ y_j\in P_2}} e^{-\frac{(l_i-l_j)^2}{\lambda}}e^\frac{||x_i-(Ry_j+y_t)||^2}{2\sigma^2}\,,
\label{eq:3D_kernel_max_label}
\end{equation}
where parameter $\lambda$ controls the sensitivity of our measure to atom labels, for example to partial charges. When $\lambda$ is large, impact of labels is negligible, which corresponds to a purely geometrical approach. When $\lambda$ is close to zero, only pairs of atoms which have exactly the same partial charge contribute to our measure. In general, the smaller $\lambda$, the greater the contribution of the atom labels to the binding pocket similarity measure.
Since the function $K_{\mbox{L}}$ does not depend on $R$ and $y_t$ in (\ref{eq:3D_kernel_max_label}), the same optimization  procedure can be used to optimize (\ref{eq:3D_kernel_max_label}) or (\ref{eq:3D_kernel_max}).

Finally, it is important to notice that the {\it sup-CK} measure of similarity can be used to compare \emph{any} set of atoms in 3D. While the primary goal of this research is to use it for comparison of binding pockets, we can also use it to compare, e.g., 3D conformations of ligands. This possibility is investigated in the experiments below.

\subsection{Related methods}   
\label{sec:related}
In this section we briefly review some of the existing methods for pocket comparison, which we compare to {\it sup-CK} in our experiments.

\noindent  \textbf{Spherical harmonic decomposition (SHD)}. \cite{Morris2005Real} proposed to model pockets by star-shapes built using the SURFNET program.  The star-shape representation is defined by a function $f(\theta,\phi)$, representing the distance from the pocket center to the pocket surface for a given ($\theta$,$\phi$). To measure the similarity of binding pockets $P_1$ and $P_2$, the corresponding functions $f_1$ and $f_2$ are first decomposed into spherical harmonics, and the  pocket similarity is then computed as the standard Euclidean metric between vectors of decomposition coefficients.
\cite{Kahraman2007Shape} presented three different variants of {\it SHD}, using only the shapes of binding pockets, the sizes of the binding pockets (keeping only the zero-th order in the spherical harmonics expansion), and their combination. We only present the results of the latter in section \ref{sec:results}, because it provided the best performance.

\noindent  \textbf{Poisson index (sup-PI)}. As we already mentioned in Section \ref{sec:convolution}, many binding pockets similarity measures are based on pocket alignment with further counting of overlapping atoms. In particular this kind of approach is used in the {\it Poisson index} model \cite{Davies2007Poisson}.   More precisely the {\it Poisson index} model is based on normalized number of overlapping atoms
$PI(P_1,P_2)=\frac{L}{\#P_1+\#P_2-L}$ where $L$ is the number of overlapping atoms, and $\#P_1$ and $\#P_2$ are the respective numbers of atoms in the two pockets. The $PI$ score may be computed for any pocket superposition method.While \cite{Davies2007Poisson} used the geometric hashing algorithm, we use in our experiments the superposition made by {\it sup-CK} method, with further superposition refining to maximize the number of overlapping atoms. 

\noindent  \textbf{Multibind}. \cite{Shulman2008MultiBind} represent pockets by pseudo-atoms labeled with physico-chemical properties. Pockets are aligned using a geometric hashing technique. This algorithm  was mainly designed for multiple alignment of binding sites, but it may be used for pairwise alignment of pockets, as was performed in this study.

\noindent  \textbf{Other simple methods}. We also consider two simple methods based on the comparison of simple binding pockets characteristics. These methods represent each pocket by an ellipsoid constructed on the basis of pocket principal axis. The first method, referred to as {\it Vol}, estimates the similarity between pockets $P_1$ and $P_2$ by the absolute value of the difference between the volumes of their corresponding ellipsoids: $Vol(P_1,P_2)=\vert Vol(P_1)-Vol(P_2)\vert$. The second method, called {\it Princ-Axis}, estimates the similarity score between pockets by  $\sum_{i=1}^3(\lambda_i^{P_1}-\lambda_i^{P_2})^2$, where $\lambda_i^{P_1}$ and $\lambda_i^{P_2}$  are the lengths of the three principle axis of pockets $P_1$ and $P_2$, respectively.

\noindent \textbf{Combination of sup-CK and Vol}.
Since volume information was found to be important by \cite{Kahraman2007Shape}, we also test a linear combination of the {\it sup-CK} and {\it Vol} methods, called {\it sup-CK-Vol}, where the coefficient of linear combination is learned as other model parameters in the double cross validation scheme. This linear combination takes advantage of the Vol method to separate very different pockets like PO4 and NAD, and of the {\it sup-CK} algorithm to allow finer discrimination.

\subsection{Performance criteria}
\label{sec:criterions}

There are various ways to measure the similarity between binding pockets, some of them were discussed in the previous section. 
To evaluate the quality of a given similarity measure, one may compare it to some "ideal" similarity measure between binding pockets, but the problem is that such measure does not exist. As an example, given two alternative measures SM1 and SM2 applied to two pockets P1 and P2 such that SM1(P1,P2)= 0.3 and SM2(P1,P2)= 0.4, there is no way to decide which one is the best because we do not have any absolute reference. The choice of the optimal measure, thus, may depend on a particular problem of interest. In the context of ligand prediction, the quality of a similarity measure can be evaluated according to its ability to regroup together pockets binding the same ligand, which can be used to predict ligands for previously unseen binding pockets. To evaluate the regrouping quality of the similarity measures, we use two different scores.\\
\textbf{AUC score}. \cite{Kahraman2007Shape} use the AUC score which is computed as follows. Let us consider a set of pockets $(P_1,\dots,P_N)$ and a similarity measure $SM$. To estimate the AUC score of a given pocket $P_*$, we rank all other pockets according to their similarity to $P_*$, $SM(P_i,P_*)$ (descending order), and we plot the ROC curve, i.e., the number of pockets binding the same ligand versus the number of pockets binding a different ligand among the top $n$ pockets, when $n$ varies from $0$ to $N$. The ranking quality of $SM$ is measured by the surface of area under the ROC curve, which defines the AUC score. An "ideal"  $SM$ function will rank all pockets binding the same ligand as $P_*$ on the top of the list, leading to an AUC score equal to $1.0$. On the contrary,  if these pockets have random positions in the ranked list, the AUC score will be equal to $0.5$ (worst possible case). Finally, to evaluate the overall AUC score of a method, we consider its mean value over all pockets.

While the AUC score represents an intuitive and natural way to evaluate the quality of similarities measures, in some situations it may fail. 
Consider the case of a dataset containing two types of pockets $L_1$ and $L_2$ (i.e. they bind two different ligands), and a similarity measure that correctly clusters pockets according to their type. If clusters are close to each other (see clusters A and C in Figure \ref{fig:ce_vs_auc}), the AUC score of pockets situated near the border (pockets $p_1$ and $p_2$ in Figure \ref{fig:ce_vs_auc}) will be low. The situation becomes even worse, if pockets binding ligand $L_1$ form several clusters, as shown in  Figure \ref{fig:ce_vs_auc}, leading to low AUC scores for almost all pockets binding ligand $L_1$. This similarity measure will have an overall poor AUC score, although it produces perfect separation of pocket types.This happens, for example, when the database contains proteins that underwent convergent evolution and 
bind the same ligand under highly different conformations. Therefore, a poor AUC score does not necessarily correspond to a poor pocket separation, and AUC scores may not be suited to evaluate the quality of similarity measures.\\
\textbf{Classification error}. These remarks lead us to employ another quality score based on classification error. To evaluate the quality of the similarity measure $SM$ we try to predict a ligand (class) for each pocket from that of its neighbors. The smaller the classification error (proportion of bad predictions), the better the similarity measure.

\begin{figure}
\centering
\includegraphics[width=8cm]{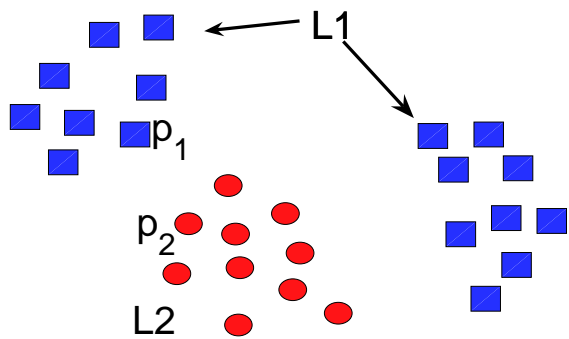}
\caption{AUC score versus classification error as an evaluation of binding pocket similarity measure. Red circles represents pockets fixing ligand $L_1$, blue squares represents pockets fixing ligand $L_2$. The AUC score does not reflect the fact of good pocket clusterization, while the classification error does.}
\label{fig:ce_vs_auc}
\end{figure}

In this work, we use a K nearest neighbors (KNN) classifier. To eva\-lua\-te the clas\-sification error, we applied a leave-one-out double cross vali\-dation methodology. Na\-me\-ly, each pocket from the dataset is considered one by one, and all other pockets are used as the trai\-ning set. Parame\-ters of the model ($k$ --- number of neighbors, $\sigma$ if we consider sup-SM method) are estimated on the training data via cross-validation techni\-que, and the class (i.e. the ligand) of the pocket under consideration is predicted using the training data and the estimated parameters of the model.

\subsection{Docking}    
\label{sec:docking}

Since docking programs may also predict ligands, we used the {\it Fred} \cite{Nicholls2005Openeye} and {\it FlexX} \cite{Rarey1996fast} programs. We chose these two programs because they are well referenced, and represent different strategies for ligand placement in the binding site. In all docking experiments, the active sites were the same as those used by the {\it sup-CK} methods.
{\it Fred} performs rigid docking of molecules. Flexibility of ligands is taken into account by using pre-calculated conformers of a molecule. These conformers are ranked according to their estimated interaction energy with the protein, which defines the docking score (\textit{chemgauss3} score) \cite{Mcgann2003Shape}. For each pocket, the predicted ligand was the most frequent molecule observed among the K first ranked molecules (K was optimized for each dataset). 

{\it FlexX} performs flexible docking of molecules by fragmentation and incremental rebuilding inside the binding site. Therefore, only one ligand conformation is required as input, and the docking results are expected to be independent from that conformation. To predict a ligand for a given pocket, we choose the molecule of best docking score. In all cases, {\it FlexX} was run using standard parameters, with formal charges, and multiple conformations for rings were computed with Corina \cite{Gasteigner1996Chemical}. 

To evaluate the performance of docking programs we can use only classification error score. Fred and Flex may be used to predict binding ligands, but they do not measure similarity between binding pockets, so we can not compute the AUC score.     
\section{Datasets}
\label{sec:datasets}
For all protein structures, the binding pockets were extracted as follows: protein atoms situated at less than $R$\AA\  of one of the ligand atoms were selected, where $R$ is considered as a model parameter and is learned in the double cross-validation scheme.  In our experiments, in most cases the optimal value of $R$ was equal to 5.3 \AA, this distance cutoff is in the range of that above which most interaction energy terms between a protein and a ligand usually become negligible. Finally, pockets are represented by 3D atom clouds with atom labeled by their partial charge, but other labels representing chemical properties such as amino-acid type could be included. Atom partial charges were attributed according to the GROMACS (FFG43a1) force field \cite{Scott99thegromos}.

We consider several benchmark datasets. The first one, referred to as the \emph{Kahraman dataset}, comprises the crystal structures of 100 proteins in complex with one of ten ligands (AMP, ATP, PO4, GLC, FAD, HEM, FMN, EST, AND, NAD). It was proposed by \cite{Kahraman2007Shape} and is described in the Supplementary Materials.
We built an extended version of the Kahraman dataset (called \emph{extended Kahraman Dataset} below), also described in the Supplementary Materials, in which we added protein structures in complex with one of the same ten ligands, leading to a total of 972 crystal structures. The added proteins present pairwise sequence identities less or equal to 30\%, to avoid potential bias by inclusion of close homologs. 

The Kahraman dataset contains only holo protein structures. However, apo structures may differ from holo structures when the latter undergo structural rearrangement upon ligand binding, a phenomenon called induced fit of the protein in order to adjust to the ligand \cite{Bosshard2001Molecular}. We tested a few examples of predictions for eight apo structures to evaluate the robustness of our method  with respect to atom positions variability. We considered 8 apo structures corresponding to proteins able to bind one ligand from the Kahraman database: 1ADE for AMP, 1B8P for NAD, 1E4F for ATP, 1OMP for GLC, 1WS9 for FAD, 2RG7 for HEM, 1X56 for PO4 and 1N05 for FMN. These proteins share less than 30\% sequence identity with any of the proteins of the extended Kahraman dataset, and had an holo structure available. The LigASite website \footnote{http://www.bigre.ulb.ac.be/Users/benoit/LigASite/} was used for this selection. The holo and apo structures of these proteins were superposed, and the coordinates of the ligand in the holo structure were used to extract the pocket in the apo structure. 

The Kaharaman dataset comprises ligands of very different sizes and chemical natures. However, the real challenge is to test methods on pockets that bind ligands of similar size. Therefore, we created a third dataset comprising 100 structures of proteins in complex with ten ligands of similar size (ten pockets per ligand). This dataset will be referred to as the \emph{Homogeneous Dataset} (HD), and is described in Supplementary Materials.

\section{Results}
\label{sec:results}

The methods were tested on two datasets (Section \ref{sec:datasets} and Supplementary Materials).
The performance of all methods is evaluated on the basis of the AUC score  and the classification error (Section \ref{sec:criterions}). The {\it sup-CK} method is compared to {\it sup-PI}, {\it SHD}, {\it Vol}, {\it Princ-Axis} and {\it MultiBind} algorithms (Section \ref{sec:related}). Among the pocket extraction methods used in the {\it SHD} approach, we considered the results corresponding to the Interact Cleft Model, which is similar to our pocket extraction method.  Results provided by the docking programs are called {\it Fred} and {\it FlexX}. 

Pocket representation is subject to extraction noise. To estimate the method performance on unnoisy systems, algorithms for pockets comparison were also employed to compare ligands (except for the {\it MultiBind} method which is designed to be employed only on proteins). 
\subsection{Kahraman Dataset}   
Results of all methods on the Kahraman Dataset are presented in Table  \ref{tab:auc_kahraman}. 
\begin{table}[h] 
\centering
\caption{Performances for all algorithms evaluated by the mean AUC scores and the mean classification errors (CE), over all pockets. We report only classification error for the  Fred and Flex docking programs, because they can not be used to evaluate similarity between binding pockets.  Column ``Pockets'' reports AUC and CE scores based on comparison of binding pockets. Column ``Ligands'' represents the same thing, but on the basis of ligands, for more explanations see text.}
\begin{minipage}{0.4\textwidth}
\begin{tabular}{|l|c|c|c|c|c|}
\hline
\multirow{2}{*}{Method}&\multicolumn{2}{|c|}{Pockets}&\multicolumn{2}{|c|}{Ligands}\\
&AUC&CE&AUC&CE\\
\hline
sup-CK&0.858$\pm$0.14 &0.36&0.964$\pm$0.006&0.04\\
$\mbox{sup-CK}_L$&0.861$\pm$0.13&0.27&---&---\\
sup-CK-Vol&0.889$\pm$0.14&0.34&0.985$\pm$0.06&0.03\\
$\mbox{sup-CK}_L$-Vol&0.895$\pm$0.12&0.26&---&---\\
Vol&0.875$\pm$0.14&{0.39}&0.897$\pm$0.13&0.30\\
Princ-Axis&0.853$\pm$0.13&{0.35}&0.938$\pm$0.10&0.16\\
sup-PI&0.815$\pm$0.13&0.42&0.927$\pm$0.09&0.05 \\
SHD \footnote{AUC scores are taken directly from \cite{Kahraman2007Shape}, CE scores are estimated from data provided by authors} & 0.770&0.39&0.920&0.07\\
MultiBind& 0.715 $\pm$0.17 &0.42&---&--- \\
\hline
Fred&---&0.47&---&--- \\
Flexx&---&0.62&---&---\\
\hline
\end{tabular}
\end{minipage}
\label{tab:auc_kahraman}
\end{table}
According to the AUC score, simple methods like {\it Vol} and {\it Princ-Axis} give surprisingly good results. The same effect was observed by \cite{Kahraman2007Shape} when they used simple measure based on comparison of pocket sizes. The AUC scores of all the new methods ({\it sup-CK}, {\it sup-CK-Vol}, with or without use of partial charges) are higher than those of {\it ICM}, {\it MultiBind}, and {\it sup-PI}, and are in the same range than those of {\it Vol} and {\it Princ-Axis}. The best results are obtained by the {\it sup-CK-Vol} algorithm, which seems to benefit from the association of volume information and of more subtle geometric details provided by the {\it sup-CK} algorithm. Another observation, is that information on atom partial charges only leads to modest improvement of the {\it sup-CK} methods. 

To evaluate the classification error, we tried to predict a ligand (a class) for each pocket using a K Nearest Neighbors classifier (see Section \ref{sec:criterions}). Note that in a ten class (10 ligands) classification problem, a random classifier would have an error of $0.9$, which represents baseline performance for all classifiers.

Table \ref{tab:auc_kahraman} shows that methods with higher AUC scores tend to have smaller classification errors, but this correlation is not strict. This indicates that the AUC score is not appropriate to compare similarity measures with respect to the problem of ligand identification, and underlines the interest of the classification approach. 
 
The {\it sup-CK} and {\it sup-CK-Vol} algorithms have lower classification errors than other methods, which means that they are well suited to the problem of ligand prediction. Interestingly, atom partial charges information significantly reduces classification errors of both methods, which was not the case for AUC scores. Addition of more information for the description of pockets may improve the quality of ligand prediction. The {\it SHD} and {\it MultiBind} methods provide reasonable prediction quality, although they do not perform as well as {\it sup-CK}.
The only difference between the {\it sup-PI} and {\it sup-CK} methods is the similarity measure used  after superposition. The {\it sup-PI} method requires to determine the number of overlapping atoms. On the contrary, the {\it sup-CK} measure is based on a weighted number of atoms having close positions score taking into account,  which probably leads to better results. 

Docking is now widely used for ligand prediction \cite{Leach2006Prediction}, and it is therefore interesting to compare its performances to those of pocket comparison methods. Table \ref{tab:auc_kahraman} shows that, on this benchmark, both docking programs do not perform as well as the {\it sup-CK} method, although {\it Fred} has better results than {\it FlexX}. Comparison of docking programs performances is beyond the scope of this paper, but it has been widely discussed that relative performances of docking programs strongly depend on the datasets  \cite{Warren2006Critical}. They were here overall modest, but both docking programs better classified pockets associated to large ligands like FAD (flavin-adenine dinucleotide) or FMN (flavin mononucleotide), and poorly those that bind smaller ligands. These results are consistent with the fact that small ligands make few interactions, leading to low docking scores.

Since {\it sup-CK} method relies on 3D atom cloud representation of protein pockets, we applied it to compare ligands using their coordinates in the protein-ligand complex structures. We also recall the performances of the {\it SHD} algorithm for ligands of this dataset. 
No method reaches an AUC score of $1.0$, or perfectly classifies the ligands (i.e. perfectly assign the correct ligand type). This indicates that ligands adopt different conformations in this dataset. However, performances of all algorithms are better for ligands than for pockets. Pockets have to be extracted from the protein structure, which introduces some noise that is absent in the case of ligands. This may explain better results, and represent the best expected performances for each method. In the case of ligand comparison, the best results are obtained with the {\it sup-CK} algorithms, although those of {\it SHD} and {\it sup-PI} are very good. The {\it Vol} and {\it Princ-Axis} methods have significantly lower results in terms of ligand classification than other methods, although their AUC scores were in the same range. Similarly, the {\it SHD} and {\it sup-PI} AUC scores are close to that of {\it Princ-Axis}, but they both perform much better in ligand classification than the latter.\\
{\bf Extension of Kahraman dataset.}\\
To evaluate the ability of the {\it sup-CK} method to improve its performance when trained on a larger dataset, we consider  an extension of Kahraman dataset consisting of 972 pockets that bind one of the 10 ligands of the original dataset (see Section \ref{sec:datasets}). Pocket comparison and ligand prediction was performed with the {\it sup-CK} method including atom partial charges. The mean AUC score and classification error were equal to $0.87$ and $0.18$. In particular, 79\% of the binding pockets of the original Kahraman dataset were correctly classified, compared to 73\% on the original dataset (see Table \ref{tab:auc_kahraman}). The results of the new method improve when trained on a larger dataset, which shows its ability to learn. The quality of predictions might again improve by including more structures available at the PDB. 

It is also interesting to study the structure of the dataset according to the metric associated to the {\it sup-CK} method. 
We performed kernel principal component analysis \cite{Schoelkopf1999Kernel} on the pockets similarity matrix of the {\it sup-CK} method (this matrix is not positive definite, but we can extract principal components associated to the largest positive eigenvalues). Figure \ref{fig:th_compl_kpca}(a) represents the projection of 972 binding pockets on the first two principal components. 
\begin{figure}[htb]
\begin{center}
\includegraphics[width=0.45\textwidth]{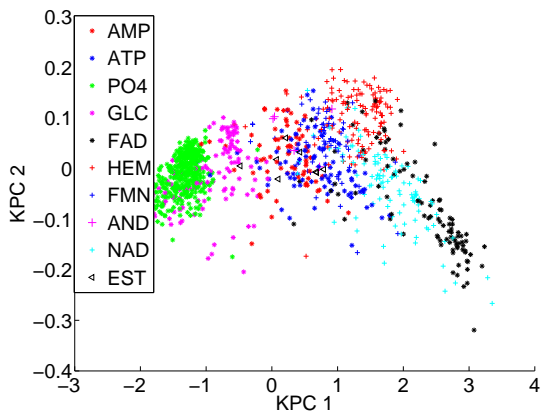}
\caption{Projection of ext-KD on the two first kernel principal components.}
\label{fig:th_compl_kpca}
\end{center}
\end{figure}
Overall, we observe a clustering of binding pockets according to their ligands, which illustrates the good performances of this method for ligand prediction. Looking into more details, we notice that the clusters of pockets that bind ATP, AMP or PO4 overlap. Indeed, proteins that binds ATP usually also bind AMP or PO4, although with different affinities. Furthermore, some pockets (for example pockets that bind glucose GLC or FAD) are found far from their main cluster, or form secondary clusters, which illustrates that pockets having different geometrical characteristics may bind the same ligand.  In the classification approach employed here, prediction of a ligand for a given pocket uses the classes of its neighbors, which allows to handle the case of pockets belonging to secondary clusters.

{\noindent \bf Prediction on apo structures}. \\
The Kahraman dataset includes protein structures in complex with a ligand which was removed, and then predicted in a "leave-one-out" procedure. However, in practice, the relevant problem will be to predict ligand for apo structures. Apo structures may differ from holo structures due to the induced fit phenomenon.   Therefore we tested the performance of our method on eight apo structures (Section \ref{sec:datasets}). The ligands for the eight considered apo pockets were predicted by the {\it sup-CK} algorithm, and the only misclassified pocket was that of 2RG7, a protein which binds HEM.
\subsection{Homogeneous dataset (HD)}		

The Kahraman dataset contains ligands of very different sizes. It is important to test methods on a benchmark containing pockets binding ligands of similar sizes. For this reason, we built the Homogeneous dataset. Table \ref{tab:auc_lapd} presents the performances of different algorithm on this dataset.

\begin{table}[htb]
\centering
\caption{Performances for all algorithms evaluated by the mean AUC scores and the mean classification errors, over all pockets.} 
\begin{tabular}{|c|c|c|c|c|}
\hline
\multirow{2}{*}{Method}&\multicolumn{2}{c|}{Pockets}&\multicolumn{2}{|c|}{Ligands}\\
&AUC&CE&AUC&CE\\
\hline
sup-CK&0.710$\pm$0.19&0.47&0.892$\pm$0.14&0.12\\
$\mbox{sup-CK}_L$&0.752$\pm$0.16&0.38&---&---\\
sup-CK-Vol&0.722$\pm$0.18&0.46&0.909$\pm$0.17&0.12\\
$\mbox{sup-CK}_L$-Vol&0.766$\pm$0.17&0.38&---&---\\
Vol&0.648$\pm$0.15&0.89&0.812$\pm$0.15&{0.54}\\
Princ-Axis&0.650$\pm$0.18&0.71&0.830$\pm$0.20&{0.28}\\
sup-PI&0.702$\pm$0.19&0.47&0.880$\pm0.14$&0.12\\
MultiBind&0.69$\pm$ 0.14&0.48&---&---\\
\hline
Fred&---&0.54&---&---\\
Flex&---&0.85&---&---\\
\hline
\end{tabular}
\label{tab:auc_lapd}
\end{table} 
Table \ref{tab:auc_lapd} shows that the performance of all algorithms are lower than on the Kahraman dataset, which illustrates that the Homogeneous dataset is a more difficult benchmark. 
The {\it Vol} and {\it Princ-Axis} display stronger degradation of performances, with AUC scores equal to 0.65, and classification errors of 89\%  and 71\%, respectively. This is due to the fact that the size information is less discriminative on this dataset.  
In terms of AUC scores, the best performance is obtained by the {\it sup-CK} and {\it sup-CK-Vol} algorithms, but volume information only provides a slight improvement of 1\%, compared to 3\% on the Kahraman dataset. On the contrary, partial charges information leads to a significant improvement of 4\% for the {\it sup-CK} and {\it sup-CK-Vol} algorithms. This shows that addition of physico-chemical information is critical for discriminating pockets of similar sizes. In terms of classification error, volume information is useless, but the use of information on partial charge leads to significant improvement of 9\%.

The same conclusions also hold for ligands comparison: performances are lower than on the Kaharaman dataset, for all methods, and degradation of  the classification errors is much stronger for the {\it Vol} and {\it Princ-Axis} methods. On this dataset, the docking programs did not perform as well as methods based on pocket comparison in terms of classification errors.
\section{Discussion}
An important characteristic of the \textit{sup-CK} algorithm is its ability to adapt to the pocket variability. Parameter $\sigma$ of the \textit{sup-CK} method controls the sensitivity of the similarity measure to atom relative displacements. The larger the variability of  pockets  binding the same ligand, the greater should be the value of $\sigma$.
Figure \ref{fig:sup_nochpart}a shows how the AUC score and classification error vary with $\sigma$ on the Homogeneous dataset. In both cases, the optimum is reached when $\sigma$ is equal to one. Note that, in our experiments (section \ref{sec:results}), we did not use the same value of $\sigma$ estimated from all pockets. For each pocket, the optimal value was estimated on the basis of the 99 training pockets to avoid overfitting to the data. However, we observed that, in most cases (90\%), $\sigma=1$ was chosen. Similarly, when information on atom partial charges is used, parameter $\lambda$ (\ref{eq:3D_kernel_max_label}) conditions the sensitivity of the method to relative values of atom charges. Figures \ref{fig:sup_nochpart}b and \ref{fig:sup_nochpart}c present the variation of AUC scores and classification error as  functions of $\sigma$ and  $\lambda$. We observe that for the AUC score, the optimum is reached when $\sigma$ equals to 2 and $\lambda$ equals to 0.25, while for the classification error optimal $\sigma$ is equal to 4.

\begin{figure*}[htbp]
 \begin{center}
\includegraphics[width=0.27\textwidth]{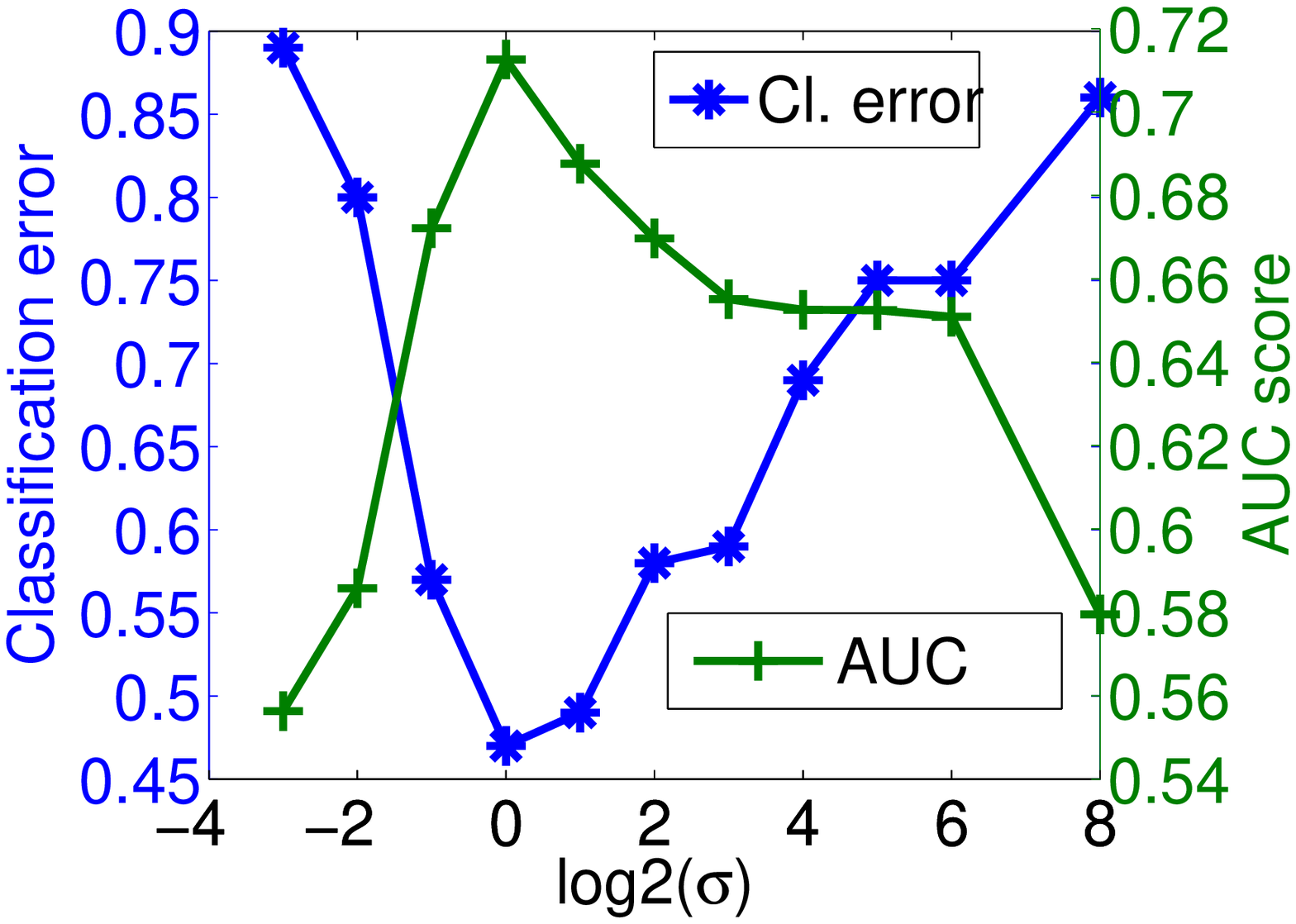} \includegraphics[width=0.27\textwidth]{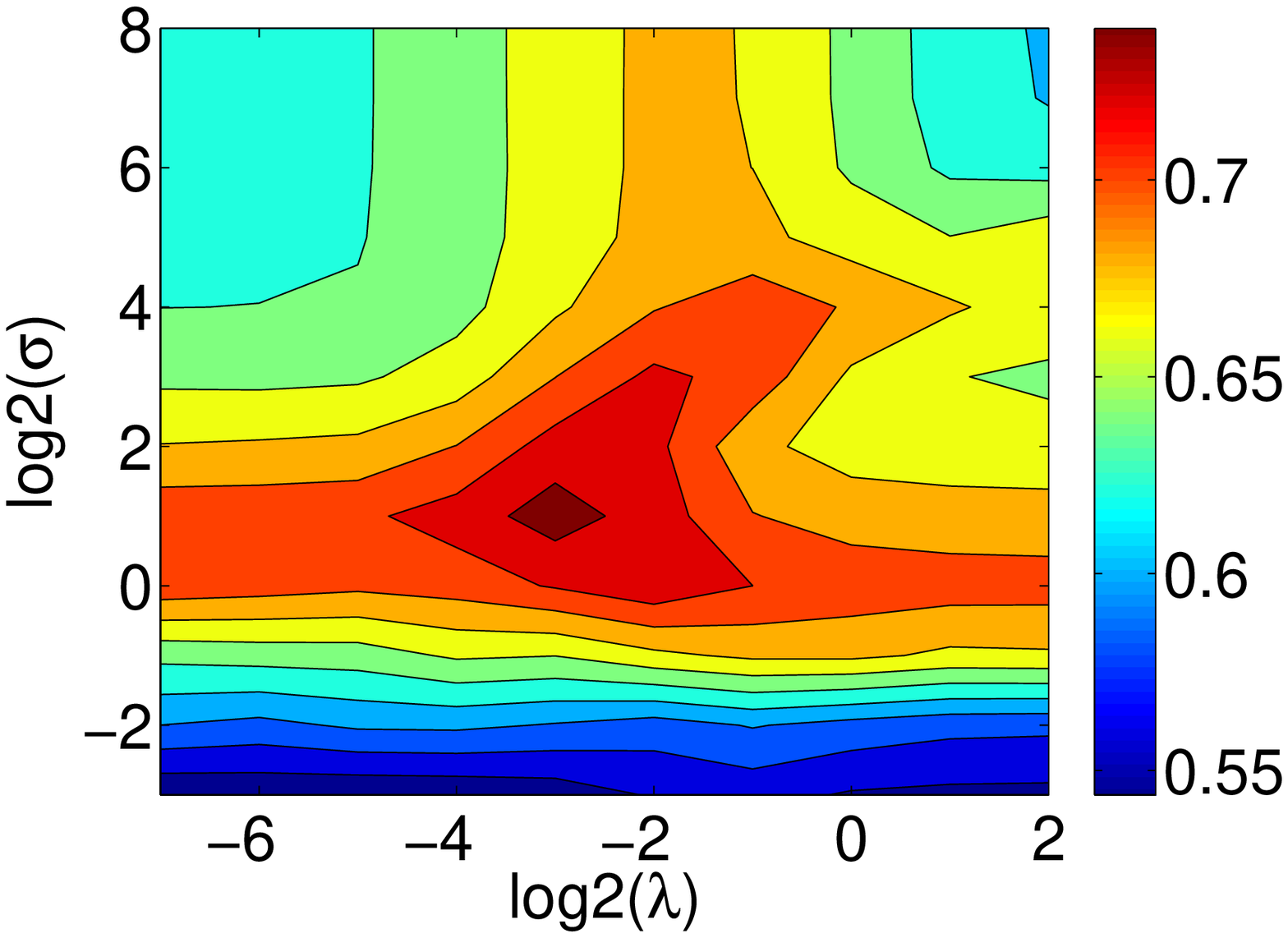} \includegraphics[width=0.27\textwidth]{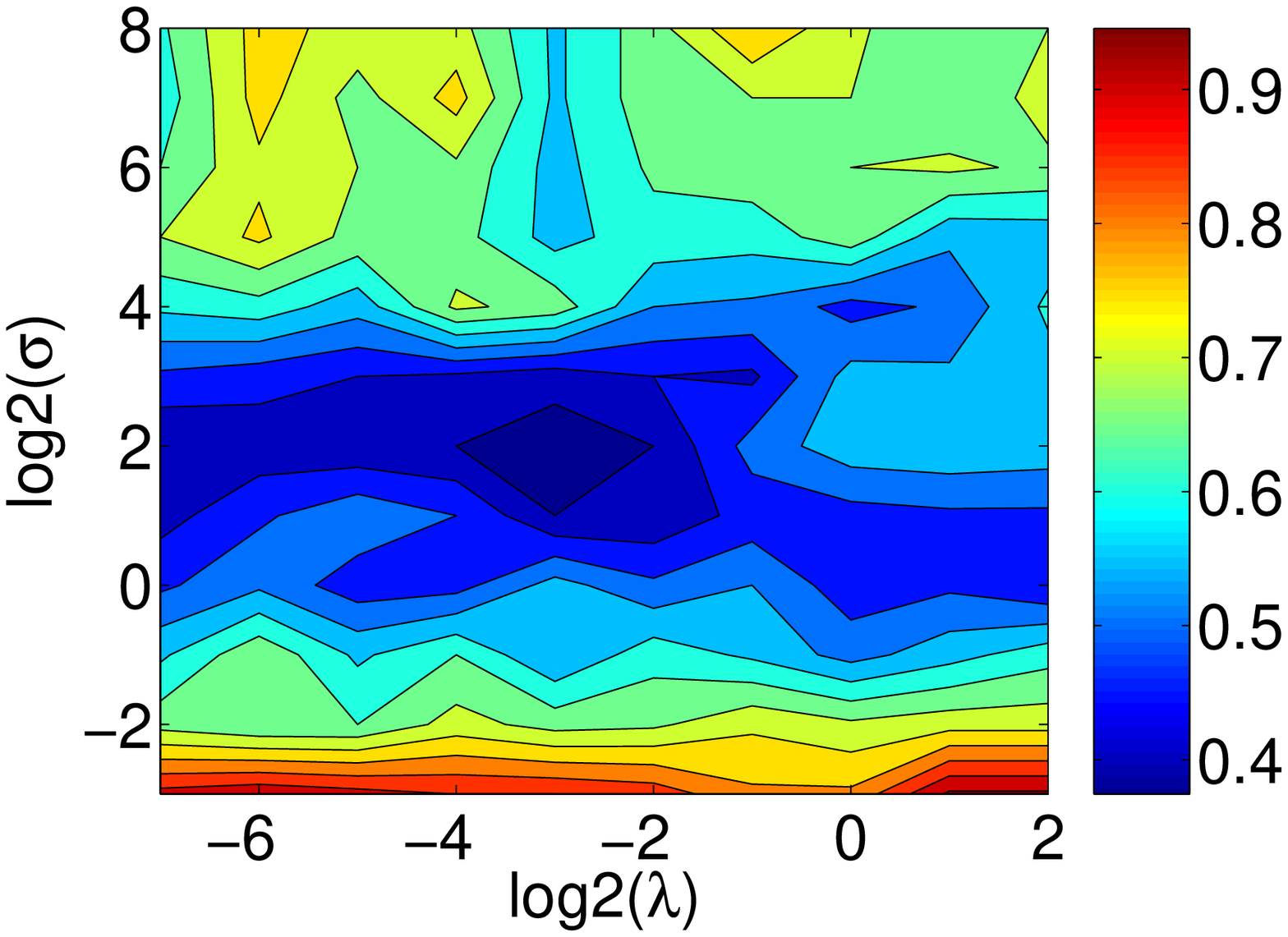}\\
{\centering (a)\qquad \qquad \qquad \qquad\qquad \qquad \qquad   (b)\qquad \qquad \qquad\qquad \qquad \qquad \qquad  (c)}
\caption{Homogeneous database. (a) AUC score and prediction error  as functions of $\sigma$ in the sup-CK method (pure geometrical version,$\lambda=\infty$), (b) AUC score  and (c) classification error as  functions of $\sigma$ and $\lambda$ when information on atom partial charges is used.}

\label{fig:sup_nochpart}
\end{center}
\end{figure*}

Figures \ref{fig:sup_examples}b and \ref{fig:sup_examples}c illustrate the optimal alignment found for two ATP binding pockets.  While this alignment was estimated on the basis of pocket atom coordinates, the bound ligands are found well aligned, which suggests a good quality of pocket  alignment. Note, that \textit{sup-CK} does not try to superpose individual atoms, but rather superposes atom sets. 

\begin{figure*}[htbp]
\begin{center}
\includegraphics[width=0.27\textwidth]{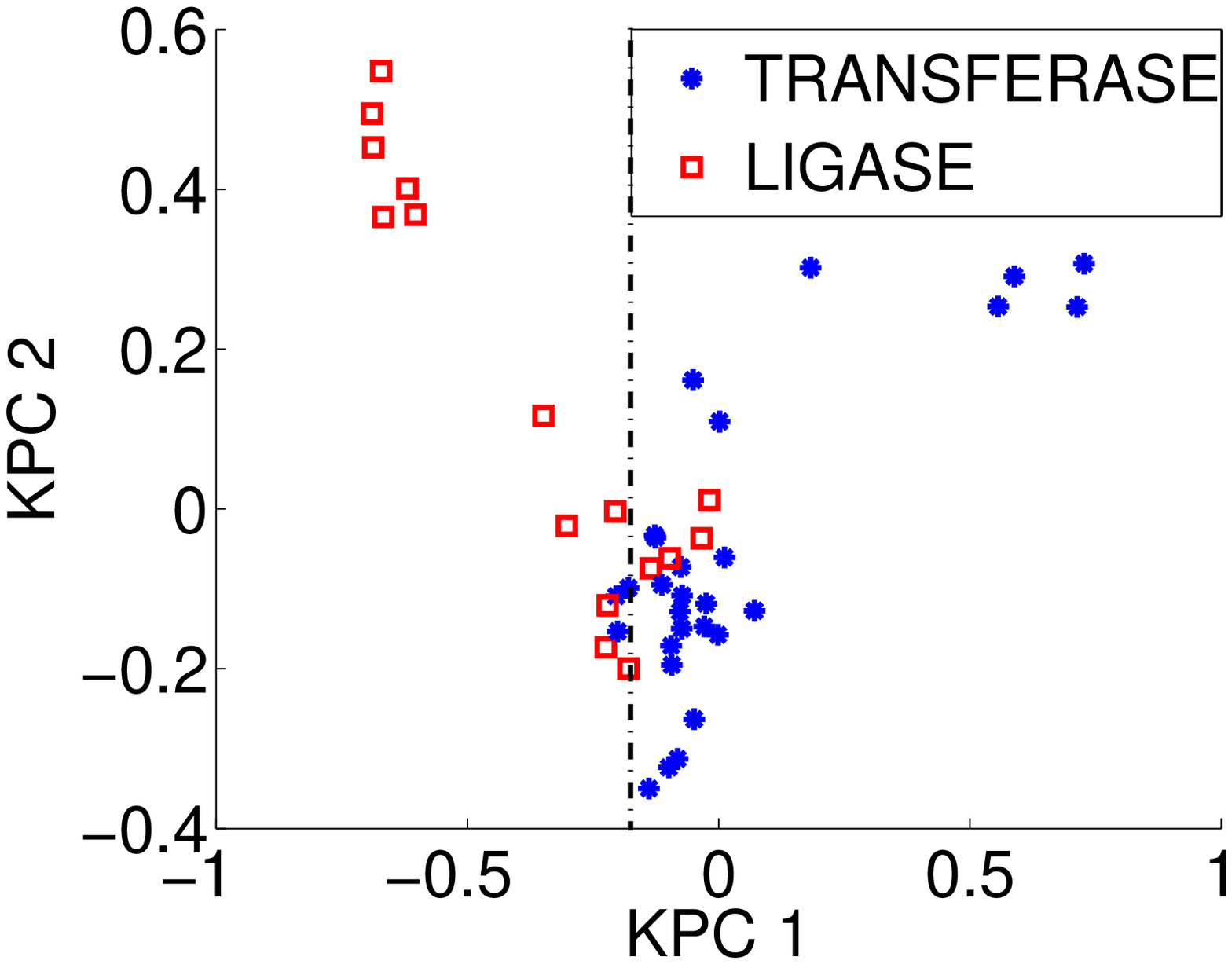}
\includegraphics[width=0.27\textwidth]{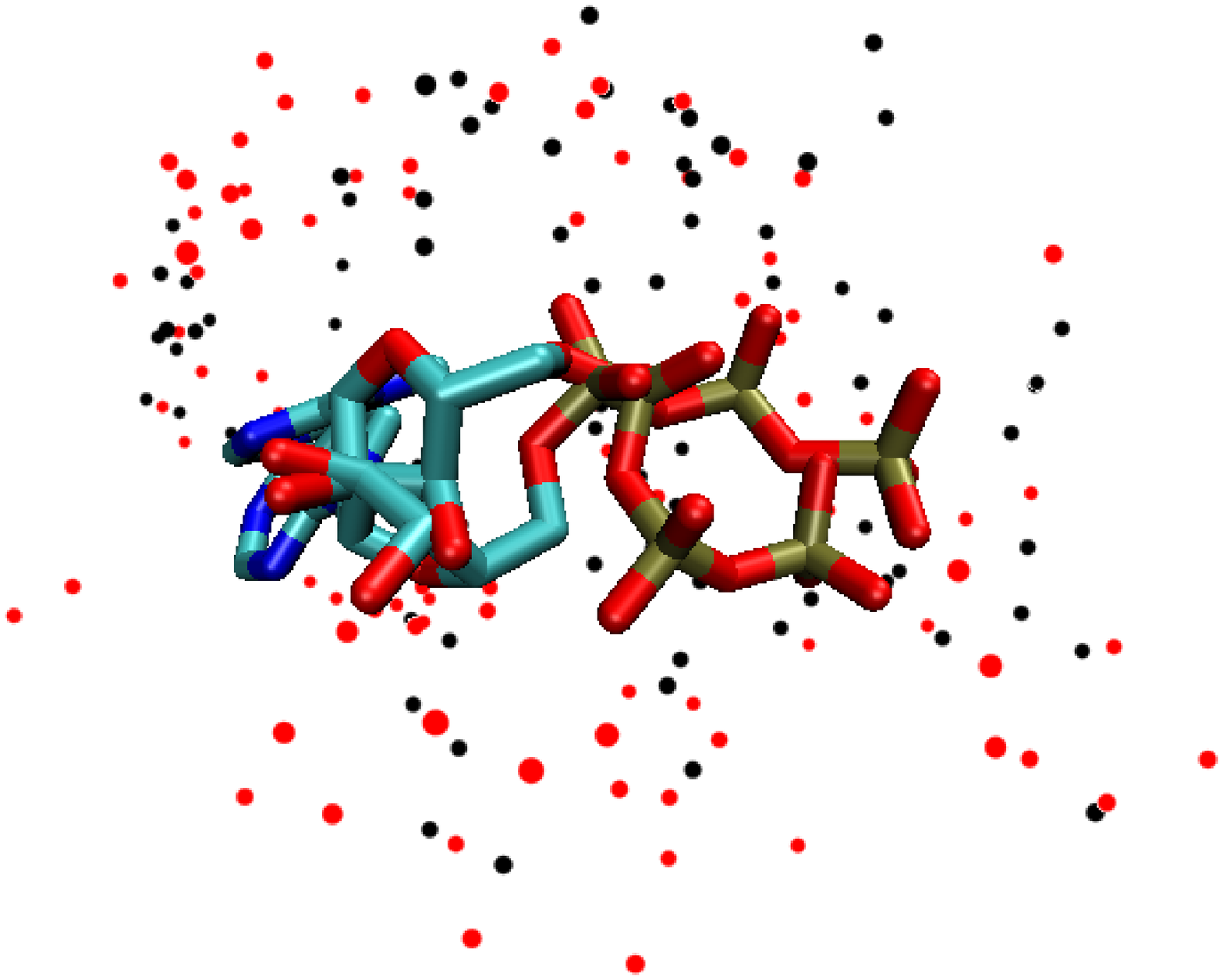} \includegraphics[width=0.27\textwidth]{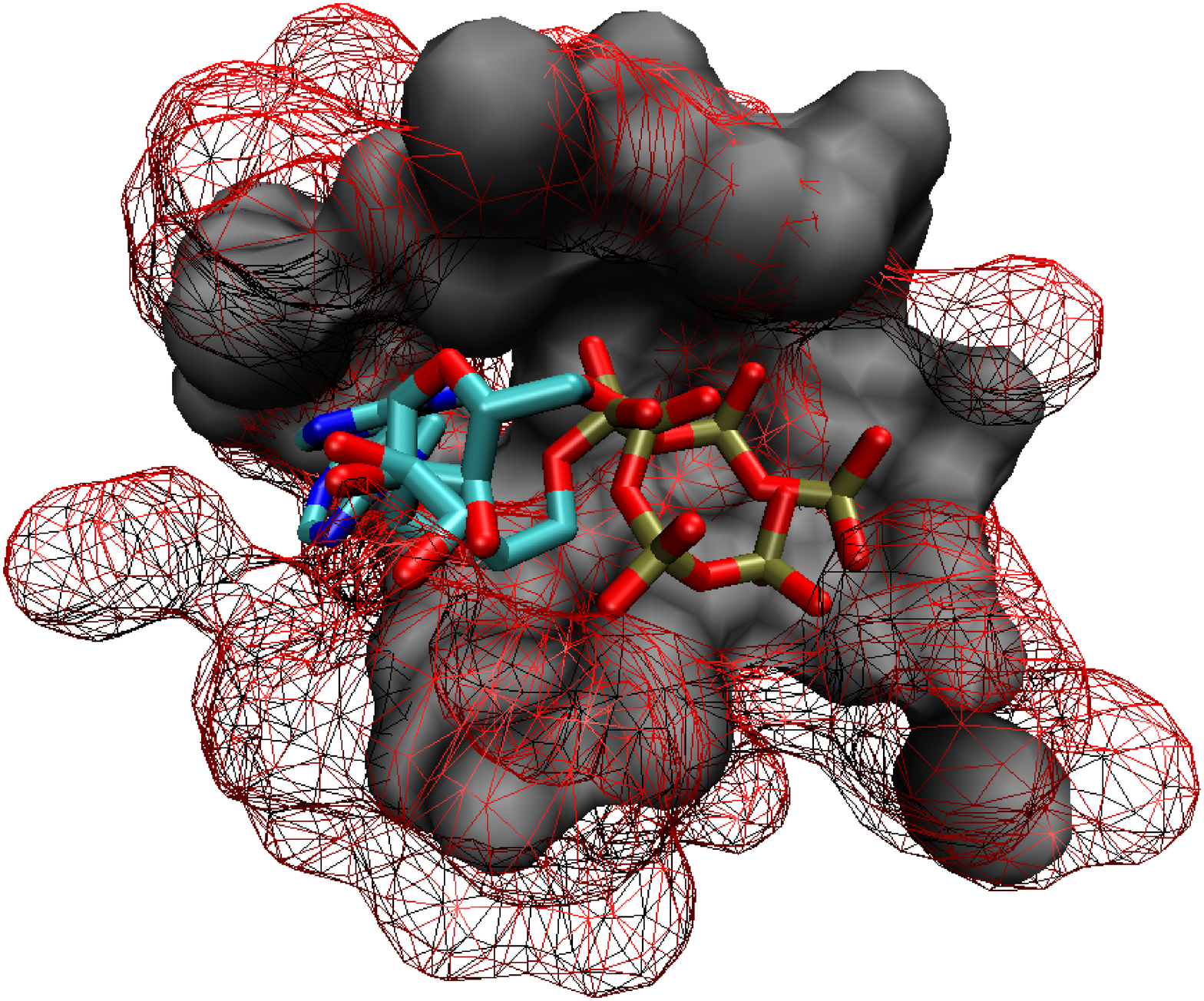}\\
{\centering (a)\qquad \qquad \qquad \qquad\qquad \qquad \qquad  (b)\qquad \qquad \qquad \qquad \qquad\qquad \qquad (c)}
\caption{(a) Projection of ATP binding pockets on the two first kernel principal components of {\it sup-CK}. (b,c) Alignment  of two ATP pockets made by {\it sup-CK}, atoms of different pockets are represented by black and red points in (b) and by black and red surfaces in (c), two ATP ligands are traced in licorice.}
\label{fig:sup_examples}
\end{center}
\end{figure*}

The running time of the \textit{sup-CK} method depends on the value of stopping criterion used in the gradient ascent method and on the number of atoms. In our experiments, the algorithm running time varied between $0.2$ and $1.3$  seconds (2.5 GHz CPU) per pocket pair. This running time is already quite reasonable to process large protein databanks, however a pre-filtering on the basis of simple pocket descriptors (like volume or size) may be quite useful in the further acceleration of the \textit{sup-CK} method.  
We defined pockets as the set of all protein atoms within 5\AA\   of a bound ligand. Similar approaches were used by \cite{Kahraman2007Shape} (Interacted Cleft Model), and similar pockets may also be retrieved by methods like \textit{Q-SiteFinder} \cite{Laurie2005Q-Site} without any information on ligand coordinates. 
 
In our experiments, docking programs ({\it FlexX} and {\it Fred}) did not perform as well for ligand prediction as most methods based on pockets similarity measure. Docking programs have many  parameters that can be tuned to particular protein-ligand systems \cite{Andersson2007Multivariate}. Fine preparation of the active site, such as assignment of amino acid protonation states, is also critical. Such tuning for each pocket is hardly automatized in large scale datasets (up to almost 1000 proteins in this study), and therefore, the performance of docking programs is underestimated.  

An important topic is the relation between methods for binding pockets comparison and algorithms in field of computer vision  for comparison of 3D shapes. A complete review of 3D shape comparison methods is  out of scope of this article, and interested readers may consult \cite{Natraj2005Three} for a detailed review. Interestingly, most of existing methods for binding pocket comparison have an analogue in the domain of computer vision. For example, methods based on real spherical harmonic expansion used in \cite{Morris2005Real} for binding pocket comparison are also discussed by \cite{Papadakis2007Efficient,Saupe20013D} in the context of general 3D shape matching.  Principles used in another popular method for matching and comparison of 3D forms, called Iterative Closest Point algorithm \cite{Zhang1992Iterative}, and its variants  are used in {\it Poisson index} and {\it MultiBind} algorithms.  Examples of approaches based on graph representation of 3D forms and graph matching methods may be found in \cite{Weskamp2007Multiple}  for binding pockets comparison, as well as in \cite{Biasotti20043D} for 3D shapes comparison. 
Nevertheless, binding pockets are not continuous shapes but discrete clouds of points. They can be transformed into 3D shapes \cite{Morris2005Real,Kahraman2007Shape}, but this transformation may be a source of noise. Moreover, a similarity measure between binding pockets should be rotationally and translationally invariant, which is not always the case in computer vision methods. However, we believe that the adaptation of appropriate methods may be very fruitful for the recognition of binding pockets. 

The prediction of protein ligands is related to the problem of predicting the protein molecular function. We analyzed the repartition of the ATP binding pockets generated by this similarity measure on the extended Kahraman dataset. Figure \ref{fig:sup_examples}a presents the projection of ATP pockets annotated as transferases or ligases, on the first two principal components of the {\it sup-CK} similarity matrix. We observed that these two families of enzymes are essentially separated. Although these are very preliminary results, they show that {\it sup-CK} method may be useful in the prediction of protein molecular functions.

The \textit{sup-CK} algorithm showed a good performance in ligand prediction for apo structures. This is an important preliminary result, in order to apply the method to real case studies, or to proteins with no known experimental structure but for which a homology model can be constructed \cite{Launay2008Homology}. 
\bibliographystyle{alpha}

\bibliography{/home/michael/These/bibli/bibli.bib}

\end{document}